\documentclass[runningheads]{llncs}

\usepackage[T1]{fontenc}
\usepackage{graphicx}
\usepackage{float}
\usepackage{hyperref}
\usepackage{adjustbox}
\newcommand{\entity}[1]{\texttt{\small\allowbreak #1}}
\newcommand{\rel}[1]{\textsl{\small #1}}

\newcommand{\triple}[3]{(\entity{#1}, \rel{#2}, \entity{#3})}

\begin{document}
\title{Rule2Text: A Framework for Generating and Evaluating Natural Language Explanations of Knowledge Graph Rules}


\author{Nasim Shirvani-Mahdavi\inst{1}\orcidID{0009-0006-2733-2242} \and \\
Chengkai Li\inst{1}\orcidID{0000-0002-1724-8278} }
\authorrunning{N. Shirvani-Mahdavi et al.}

\institute{University of Texas at Arlington, Arlington, Texas 76019, USA 
\email{nasim.shirvanimahdavi2@mavs.uta.edu}\\
\email{cli@uta.edu}}

\maketitle              
\begin{abstract}
Knowledge graphs (KGs) can be enhanced through rule mining; however, the resulting logical rules are often difficult for humans to interpret due to their inherent complexity and the idiosyncratic labeling conventions of individual KGs. This work presents Rule2Text, a comprehensive framework that leverages large language models (LLMs) to generate natural language explanations for mined logical rules, thereby improving KG accessibility and usability. We conduct extensive experiments using multiple datasets, including Freebase variants (FB-CVT-REV, FB+CVT-REV, and FB15k-237) as well as the ogbl-biokg dataset, with rules mined using AMIE 3.5.1. We systematically evaluate several LLMs across a comprehensive range of prompting strategies, including zero-shot, few-shot, variable type incorporation, and Chain-of-Thought reasoning. To systematically assess models' performance, we conduct a human evaluation of generated explanations on correctness and clarity. To address evaluation scalability, we develop and validate an LLM-as-a-judge framework that demonstrates strong agreement with human evaluators. Leveraging the best-performing model (Gemini 2.0 Flash), LLM judge, and human-in-the-loop feedback, we construct high-quality ground truth datasets, which we use to fine-tune the open-source Zephyr model. Our results demonstrate significant improvements in explanation quality after fine-tuning, with particularly strong gains in the domain-specific dataset. Additionally, we integrate a type inference module to support KGs lacking explicit type information. All code and data are publicly available at \sloppy \href{https://github.com/idirlab/KGRule2NL}{https://github.com/idirlab/KGRule2NL}.
\end{abstract}

\section{Introduction}\label{sec:intro}
Knowledge graphs (KGs) encode factual information as triples of the form (subject \entity{s}, predicate \rel{p}, object \entity{o}). They are integral to a wide range of artificial intelligence tasks and applications~\cite{ji2021survey}. Although large-scale KGs (e.g., Freebase~\cite{bollacker2008freebase} and Wikidata~\cite{vrandevcic2014wikidata}) contain a vast number of triples, they are often incomplete, which adversely affects their usefulness in downstream applications. However, KGs often hold sufficient information to infer new facts~\cite{amieplus,shirvani2025large,ghanbarizadeh2026efficient}. 
For example, if a KG indicates that a certain woman is the mother of a child, it is quite likely that her husband is the child’s father. 

Identifying logical rules—formal expressions that capture patterns and relationships such as the example above—serves multiple critical functions. These rules enable the inference of missing facts, facilitate error detection in existing data, reveal underlying patterns in large-scale datasets, and provide explanatory frameworks for automated predictions~\cite{nakashole2012query,amieplus}. AMIE~\cite{galarraga2013amie,betz2023pyclause} and AnyBURL~\cite{meilicke2024anytime} are among such rule learning systems that derive Horn clauses for symbolic reasoning in KG completion.

Despite their utility, logical rules present significant interpretability challenges for humans, particularly non-experts who must work with KG-based systems in domains such as healthcare and scientific research. The difficulty stems from several factors: the abstract nature and the complexity of logical structures, and the often opaque entity and relation labeling conventions employed across different KGs. For instance, as explained \sloppy in~\cite{shirvani2023comprehensive}, label of predicates in the Freebase dataset follow the format /[domain]/[type]/[label] (e.g., 
\rel{/american\_football/player\_rushing\_statistics/team}). Without proper background knowledge of these conventions, interpreting and validating logical rules becomes prohibitively difficult. Natural language explanations of logical rules offer a promising solution to bridge this interpretability gap. While predefined templates could generate such explanations, this approach lacks scalability for the thousands of rules typically extracted from large KGs. Recent advances in large language models (LLMs) present an opportunity to address this challenge through their demonstrated capabilities in natural language generation and logical reasoning.

We present Rule2Text, a complete framework addressing this challenge with several key contributions: (1) extensive experiments across diverse domains from general knowledge to specialized biomedical datasets; (2) development and validation of an LLM-as-a-judge~\cite{zheng2023judging} evaluation framework enabling scalable quality assessment; (3) creation of high-quality ground truth datasets for both general and domain-specific contexts; (4) successful fine-tuning of open-source models using our generated datasets; and (5) integration of type inference capabilities for KGs lacking explicit entity type information. To our knowledge, this is the first comprehensive study examining the effectiveness of LLMs for generating natural language explanations of knowledge graph rules.

Our findings demonstrate that combining Chain-of-Thought~\cite{wei2022chain} prompting with variable type information yields substantial improvements in explanation quality, with Gemini 2.0 Flash achieving the highest correctness and clarity scores on human evaluation. Our LLM-as-a-judge framework shows strong agreement with human annotators, enabling scalable evaluation. Most notably, fine-tuning open-source models on our generated datasets produces dramatic improvements in content coverage and semantic similarity. 

The remainder of this paper is organized as follows: Section~\ref{sec:related} reviews related work in natural language generation from logical forms. Section~\ref{sec:motiv-back} provides motivation and background on rule mining. Section~\ref{sec:methodology} details our methodology including prompt engineering, type extraction, and dataset creation. Section~\ref{sec:evaluation} describes our evaluation framework and LLM-as-a-judge design. Section~\ref{sec:experiments} presents experimental setup, Section~\ref{sec:results} reports results across all experiments, and Section~\ref{sec:conclusion} concludes the work.

\vspace{-.2 cm}
\section{Related Work}\label{sec:related}
Prior work in natural language generation from logical forms, such as Logic2Text~\cite{chen2020logic2text} and SLEtoNL~\cite{wu2023generating}, generates high-fidelity text from structured tables and sequential logic. While effective for their specific domains, these models are limited for explaining knowledge graph rules. They rely on input structures that are fundamentally different from the Horn clauses we use. Furthermore, they prioritize fluent summaries over pedagogical explanations and neglect the crucial semantic roles of entities, types, and relations within knowledge graphs.

Another relevant area is the encoding and translation of natural rules~\cite{clark2020transformers,aesoy2023rule,yang2023logical,kiashemshaki2025simulating}, which converts natural language rule expressions into a formal logical format. This work aims to acquire rules from human expertise, whereas our approach focuses on interpreting and explaining existing rules. Similarly, Chain of Logic~\cite{servantez2024chain} improves how large language models apply compositional rules to factual scenarios. However, this approach also assumes the rules are already available and ready for reasoning, which differs from our goal of providing explanations for them.

In the broader context of KG-to-text generation, Shi et al.~\cite{shi2023hallucination} tackle the challenge of generating natural language descriptions from KG triples while mitigating hallucinations in large-scale, open-domain settings. While their work shares the goal of converting structured knowledge into natural language, it focuses on factual triple descriptions rather than rule explanations.
\vspace{-.2 cm}
\section{Motivation and Background}\label{sec:motiv-back}
\subsection{Motivation}\label{sec:motivation}
The proliferation of complex AI systems across critical applications has intensified demands for algorithmic transparency and explainability. This need is particularly acute in high-stakes domains such as healthcare, where stakeholders require clear justification for automated decisions. KG technologies represent one such class of systems where explainability is paramount.
Logical rules extracted from KGs serve multiple crucial functions: they enable inference of missing facts with high probability, facilitate error detection in existing data, reveal underlying patterns in large-scale datasets, and provide explanatory frameworks for specific predictions~\cite{nakashole2012query,amieplus}. These capabilities make rule-based reasoning an essential component of KG completion and quality assurance workflows.

However, a significant barrier exists between the formal representation of these rules and human comprehension. This interpretability gap stems from several factors: the abstract nature of logical structures, the complexity of multi-atom rules, and the often opaque entity and relation labeling conventions employed in different knowledge graphs. For instance, as explained \sloppy in~\cite{shirvani2023comprehensive}, label of predicates in the Freebase dataset follow the format \rel{/[domain]/[type]/[label]} (e.g., 
\rel{/american\_football/player\_rushing\_statistics/team}).
This opacity presents challenges not only for domain experts but also for technical practitioners who must develop, debug, and maintain large-scale knowledge graphs. The resulting comprehension burden limits the practical utility of rule-based knowledge graph systems and impedes their adoption in scenarios requiring human oversight or collaboration. 

\subsection{Rule Mining Algorithm}\label{sec:rule-alg}
We employed AMIE 3.5.1, a well-established rule learning system in its latest version, released in 2024, due to its comprehensive metrics for rule evaluation as well as its proven compatibility with our chosen benchmark datasets. In AMIE, a rule has a body (antecedent) and a head (consequent), represented as $B_1 \land B_2 \land \ldots \land B_n \Rightarrow H$
, or in simplified form \(\overrightarrow{B} \Rightarrow H\).  
The body consists of multiple \textit{atoms} $B_1$, $\ldots$, $B_n$, and the head $H$ itself is also an atom. In an atom \rel{r}(\entity{h},\entity{t}), which is another representation of a triple \triple{h}{r}{t}, the subject and/or the object are variables to be instantiated. The prediction of the head can be carried out when all the body atoms can be instantiated in the KG. 

For instance, consider the following simple rule extracted from the Freebase dataset: \triple{?a}{/spaceflight/bipropellant\_rocket\_engine/oxidizer}{Hydrogen peroxide} $\Rightarrow$ \triple{?a}{/spaceflight/rocket\_engine/manufactured\_by}{NPO Energomash}. This rule consists of a single atom in the body and a single atom in the head. The entities \entity{Hydrogen peroxide} and \entity{NPO Energomash} are constant entities, while \entity{?a} is a variable entity that can be instantiated with entities that satisfy the rule. For instance, in the following instantiation: \triple{RD-161P}{/spaceflight/bipropellant\_rocket\_engine/oxidizer}{Hydrogen peroxide} $\Rightarrow$ \triple{RD-161P}{/spaceflight/rocket\_engine/manufactured\_by}{NPO Energomash}, \entity{?a} is replaced by \entity{RD-161P}, which is a rocket engine. 

In AMIE, the concept of \textit{support} quantifies the amount of evidence (i.e., correct predictions) for each rule in the data. It is defined as the number of distinct (subject, object) pairs in the head of all valid instantiations of the rule in the KG. The concept of \textit{head coverage}, a proportional version of \textit{support},  is the fraction of \textit{support} over the number of facts in relation \rel{r}, where \rel{r} is the relation in the head atom. The \textit{standard confidence} of a rule is the fraction of \textit{support} over the number of instantiations of the rule body.  
\section{Methodology}\label{sec:methodology}
Rule2Text is an LLM-powered framework for generating natural language explanations of logical rules extracted from KGs. The framework employs a modular architecture where the rule mining algorithm serves as a pluggable component, enabling integration with various KG rule extraction methods. Since the focus of our study is on KG completion, we utilize the AMIE algorithm to extract Horn rules as our rule mining approach.
Rule2Text addresses two core challenges in rule-to-text generation. First, our preliminary results~\cite{shirvani2025rule2text} show that LLMs frequently exhibit confusion regarding variable entity types within logical rules, necessitating a dedicated type inference module.  Second, the absence of suitable ground-truth datasets for training rule explanation models requires a methodology for constructing high-quality training data. To address this latter challenge, we propose a ground-truth data generation approach detailed in Section~\ref{sec:groundtruth}.
The framework additionally provides comprehensive evaluation mechanisms for assessing explanation quality and establishes a methodology for fine-tuning LLMs for the rule explanation task.
\vspace{-0.5 cm}

\begin{figure*}
    \centering
\includegraphics[width=0.75\linewidth]{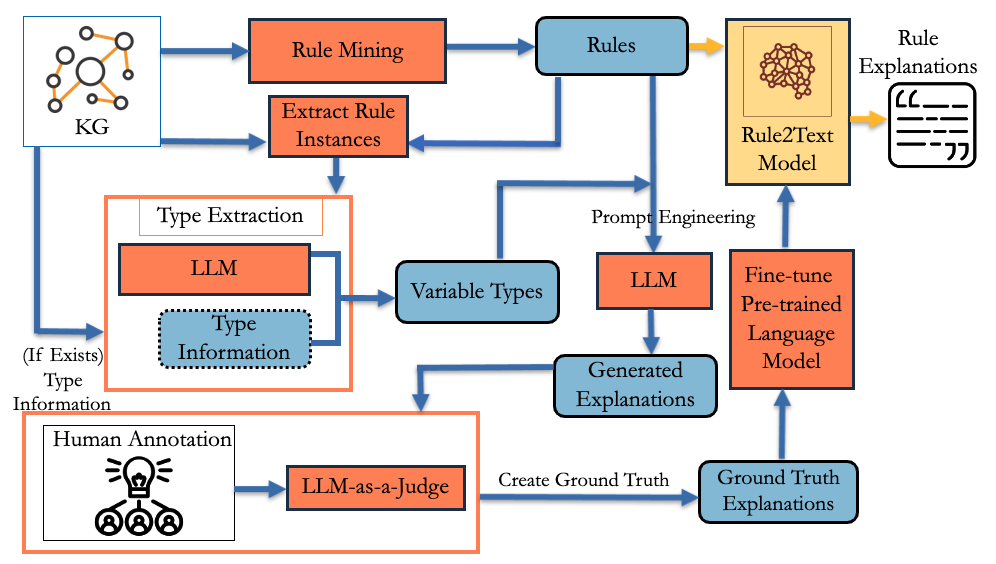}
\vspace{-.2 cm}
    \caption{Rule2Text Framework}
    \label{fig:framework}
\end{figure*}

\vspace{-0.9 cm}
\subsection{Prompt Engineering}\label{sec:prompt}
To generate natural language explanations for logical rules, we conducted prompt engineering experiments in three phases. All experimental materials—scripts, prompts, rules, generated explanations, and annotated data—are available in our GitHub repository, with preliminary results reported in our short paper~\cite{shirvani2025rule2text}. Across all experiments, we provided background knowledge to enhance model understanding of dataset syntax and labels, including predicate formats described in Section~\ref{sec:motivation}. This contextual information proved essential for handling rules with concatenated relations, detailed in Section~\ref{sec:data}, where lengthy, multi-component labels can confuse the models. 

\textbf{Zero-Shot \& Few-Shot Prompting}\hspace{3 mm} In the first phase, we compared zero-shot and few-shot prompting strategies using rules from FB15k-237~\cite{bordes2013translating}, a subset of Freebase. Our objective was to assess how in-context examples affect explanation quality and establish baseline performance. The few-shot prompts included two (rule, explanation) pairs as examples. For this phase and phase 2, we used OpenAI's GPT-3.5 Turbo~\cite{openai-gpt3-5-turbo} for its optimal balance of performance, efficiency, and cost-effectiveness.
We selected 100 rules with the highest head coverage for human evaluation, spanning diverse domains from music and media to medicine and space. To ensure annotation quality and reduce subjectivity, three annotators independently evaluated each rule. Annotators were provided with both the logical rule and a concrete instantiation to aid comprehension, along with explanations generated using zero-shot and few-shot approaches. The evaluation metrics are detailed in Section~\ref{sec:eval}. As reported in Section~\ref{sec:results}, few-shot prompting showed no significant improvement over the zero-shot baseline.

\textbf{Incorporating Variable Entity Type in the Prompt}\hspace{3 mm} Analysis of the generated explanations revealed persistent limitations in the model's ability to identify variable entity types, leading us to adopt integration of these types in the prompt in phase 2. For instance, in the rule \entity{?b} \rel{/time/event/instance\_of\_recurring\_event} \entity{World Series} $\Rightarrow$ \entity{World Series} \rel{/sports/sports\_championship/events} \entity{?b}, \entity{World Series}, the variable \entity{?b}'s type is \entity{/sports/sports\_championship\_event}.
For this phase, three annotators annotated 100 rules, rules with the highest head coverage from two large-scale Freebase datasets, 50 top rules from the FB-CVT-REV~\cite{shirvani2023comprehensive} dataset, and 50 from the FB+CVT-REV dataset.  Our findings, discussed in Section~\ref{sec:results}, show that providing variable type information significantly improved the model's performance in generating accurate explanations.

\textbf{Chain-of-Thought Prompting \& Comparing Models}\hspace{3 mm} In phase 3, building on the strong impact of incorporating variable entity types into the prompts, we further leveraged the reasoning capabilities enabled by CoT prompting. This prompt guides the model through five reasoning steps, detailed in our short paper~\cite{shirvani2025rule2text}. In this phase, we expanded our evaluation to include two additional models, GPT-4o mini~\cite{openai-gpt4o-mini} and Gemini 2.0 Flash~\cite{google-gemini-2-0-flash}, alongside GPT-3.5 Turbo. These models were selected to provide a balanced comparison in terms of performance, efficiency, and cost-effectiveness. Three annotators evaluated new explanations, generated via CoT prompting by the three models, for the same set of rules used in phase 2. As discussed in Section~\ref{sec:results}, GPT-3.5 Turbo shows improved performance compared to phase 2, while Gemini 2.0 Flash achieves the highest overall performance, followed by GPT-4o mini.

\subsection{Variable Entity Type Extraction}\label{sec:type}
As demonstrated in Section~\ref{sec:prompt}, determining the types of variable entities poses a significant challenge for language models. Therefore, a crucial step in our framework is providing the LLM with these entity types. Some KGs, such as Freebase and Wikidata, include type systems that facilitate this process. In Freebase datasets, entities can belong to multiple types. Given a property (edge) type and its instances, there exists an \emph{approximate} function that maps from the edge type to a type shared by all subjects in the edge instances, and similarly for objects~\cite{shirvani2023comprehensive}. Thus, knowing the property label allows us to infer the types of entities participating in that relation.
For example, consider the rule from Section~\ref{sec:prompt}: \entity{?b} \rel{/time/event/instance\_of\_recurring\_event} \entity{World Series}. While \entity{?b} could generally be of type \entity{/sports/sports\_championship\_event} or \entity{/time/event}, within this specific relation, its type is \entity{/sports/sports\_championship\_event}.

However, not all datasets include explicit type information. Therefore, we need an approach to infer the types of variable entities. This can be accomplished similarly to how humans learn what types of entities can instantiate rules; by providing several examples to the model. To this end, we extract three random instances of each rule and provide them to the LLM, asking it to infer entity types based on the rule context and property labels. Our results, discussed in Section~\ref{sec:results}, demonstrate that the LLM performed this task effectively overall. The scripts for extracting rule instances and performing type inference are available in our GitHub repository.

\subsection{Dataset Creation}\label{sec:groundtruth}
In the prompt engineering step, we leveraged proprietary LLMs with strong reasoning capabilities, such as Gemini 2.0 Flash. However, for various reasons, including cost efficiency and the need to run models locally due to data sensitivity concerns (e.g., healthcare data), we can fine-tune open-source models such as Zephyr~\cite{tunstall2023zephyr7bbeta}, which is known for its instruction-following abilities. However, the lack of ground-truth data presents a limitation. Therefore, we developed an approach for ground-truth data construction for fine-tuning.

One approach would be to hire human annotators to generate natural language explanations for logical rules. However, this approach has several limitations. First, annotators must be experts in the field and familiar with logical forms. Second, we would need to provide training for each dataset, as labeling syntax varies across different KGs. Furthermore, this approach can be costly and laborious.

To simplify and accelerate this process, we can leverage the best-performing model, Gemini 2.0 Flash. This model demonstrates reasonable correctness and clarity, as detailed in Section~\ref{sec:results}. We employ it to generate natural language explanations for the desired number of ground-truth data instances. Human annotators then evaluate these generated explanations. If the explanations are not perfectly correct, annotators modify them rather than writing complete explanations from scratch.

This approach allows us to generate ground-truth data using a relatively strong model. The resulting dataset is significantly smaller than the complete set of rules extracted from the KG. We can then use this data to fine-tune the open-source model. However, the limitations mentioned above still exist. Although the task becomes easier and faster for human annotators, this approach does not solve the problem fundamentally.

To facilitate this further, a potential solution lies in leveraging an LLM-as-a-judge~\cite{zheng2023judging} as an evaluator. If a reliably fair and consistent judge model can be designed, it becomes possible to automatically evaluate rule-explanation pairs. The judge can assess these pairs, and those receiving high correctness scores can be treated as pseudo-ground truth for fine-tuning smaller open-source models. Additionally, low-scoring examples can be analyzed by human annotators to identify patterns that challenge even well-performing models like Gemini 2.0 Flash. Only these challenging instances would need human evaluation and modification before inclusion in the ground truth data. The design of our LLM-as-a-judge is discussed in Section~\ref{sec:eval-judge}. 
\section{Evaluation}\label{sec:evaluation}
\subsection{Evaluation Metrics}\label{sec:eval} 
To evaluate the generated explanations in experiments lacking reference data, we employed the following metrics for human and automatic evaluation. An upward arrow beside a metric indicates that higher values are better for that evaluation criterion. Conversely, a downward arrow indicates that lower values are preferred.

\textit{Correctness$^\uparrow$}: Evaluation of the explanation's accuracy on a scale from 1 (completely incorrect) to 5 (fully correct). Correctness refers to the explanation's inclusion of all components of the rule, presented in the exact logical flow specified by the rule. This metric does not measure the readability or comprehensibility of the explanation.

\textit{Clarity$^\uparrow$}: Evaluation of the explanation on a scale from 1 (very unclear) to 5 (very clear). Clarity refers to the ease with which the explanation can be understood and how naturally it reads. This metric exclusively assesses the explanation, independent of the correctness of the underlying rule.

\textit{Number of missed entities$^\downarrow$}: 
The number of entities present in the rule but not stated in the explanation.

\textit{Number of missed relations$^\downarrow$}: 
The number of relations (i.e., predicates) present in the rule but not stated in the explanation.

\textit{Number of hallucinated entities$^\downarrow$}:
The number of entities absent from the rule but incorrectly stated in the explanation.

\textit{Number of hallucinated relations$^\downarrow$}: 
The number of relations absent from the rule but incorrectly stated in the explanation.

\textit{Rule logicalness$^\uparrow$}: Although the meaningfulness of a rule is not directly related to the generation of explanations, we asked the annotators to rate the rules on a scale from 1 (not logically sound) to 2 (moderately logical), and 3 (logically sound). This metric exclusively evaluates the rule itself, without considering the explanation. 

\textit{Perplexity$^\downarrow$}: Given the absence of reference sentences for comparison with the explanations in some of the experiments, as our automatic evaluation metric, we computed perplexity~\cite{jurafsky2009speech} using GPT-2~\cite{radford2019language}. Perplexity measures how well a language model predicts a sequence of words, with lower values indicating more predictable and naturally flowing text. We use this metric because it provides an automatic assessment of the linguistic quality and naturalness of generated explanations without requiring reference data. While it is a useful measure of the model's fluency and coherence, it is not an indication of the correctness of the explanations.

\subsection{Human Annotation}\label{sec:eval-annotator}
To mitigate subjectivity and bias in the human evaluation process, we employed three expert annotators, ensuring that each data instance was independently annotated by all three. Each annotator possessed prior experience with logical forms and received training on the syntax of the knowledge graphs used in this study, as well as the evaluation metrics and assessment criteria. To further minimize potential bias, the annotators were blinded to the source model of each explanation. Moreover, a subset of rule-explanation pairs was randomly selected for discussion among the annotators to review the rationale behind their scoring decisions. This process helped identify occasional human errors and facilitated the refinement of annotations, thereby improving the overall quality and consistency of the evaluation scores.

\subsection{LLM-as-a-Judge}\label{sec:eval-judge}
LLMs are increasingly employed for performing evaluation tasks. The LLM-as-a-judge paradigm~\cite{chiang2023can,zheng2023judging} presents several advantages over traditional human-centric evaluation methods. First, it offers significant scalability by reducing reliance on human annotators, thereby enabling the development of large-scale benchmarks and accelerating experimental iterations. Second, it enhances interpretability, as LLM-based evaluators can produce both quantitative scores and accompanying rationales that explain their judgments. 

Our LLM-as-a-judge evaluation design incorporates several key considerations to ensure reliability and reduce bias. First, the evaluation instructions provided to the model were carefully constructed to be unambiguous, with a strong emphasis on distinguishing evaluation from generation. Specifically, the prompt explicitly instructs the model to assess the quality of the given explanation, not to regenerate it. This distinction is reinforced through structured CoT reasoning steps that include objective, verifiable questions. For example: "Do all variable entities stated in the rule appear in the explanation?"—a question with a binary (yes/no) answer. A follow-up prompt then asks: "If your answer is no, which variable entities are missing from the explanation?" These types of questions are designed to compel the model to compare content rather than generate a new explanation, which is critical. As generating its own explanation may lead the model to treat that output as a presumed ground truth—even if it is incorrect in cases—thereby introducing bias into its evaluation.

Second, the scoring rubric was made explicit to the model, with clearly defined criteria for each score level. For example, the prompt includes explanations of what a score of 5 or 4 signifies in concrete terms. To enhance in-context understanding, we employed few-shot prompting, presenting the model with three exemplar explanations annotated with corresponding scores to guide its interpretation. Finally, to assess the consistency of the model’s evaluation behavior, we performed a reliability check by prompting the model to score each explanation three times. This allowed us to measure intra-model consistency and identify potential variability in its assessments.

One notable form of bias associated with LLM-based evaluation is self-enhancement bias\cite{zheng2023judging}, wherein language models tend to favor responses they have generated themselves. To assess the presence of this bias in our setup, we conducted a preliminary study~\cite{shirvani2025rule2text}, showing unlike GPT-4o Mini, the best-performing model, Gemini 2.0 Flash, did not exhibit self-enhancement bias under the given experimental conditions. Based on these findings, we selected Gemini 2.0 Flash as the LLM judge for all subsequent evaluations.

This design can be enhanced by incorporating a subset of human-annotated data as ground truth. This data enables the evaluation of the judge’s performance and facilitates the refinement of prompt engineering strategies to improve their effectiveness. To this end, we employed 100 data instances from two large-scale datasets, FB-CVT-REV and FB+CVT-REV, whose generated explanations were evaluated by human annotators during phases 2 and 3 of prompt engineering. Section~\ref{sec:results} reports the agreement between the human judgments and those produced by the LLM judge. Overall, the level of agreement suggests a promising direction for adopting LLM-based evaluation in this context.

\vspace{-.6 cm}
\begin{table}[ht]
\centering
\setlength{\tabcolsep}{9pt}
\caption{Statistics of the datasets}\label{tab:datastat}
\vspace{-.2 cm}
\scalebox{0.95}{
\begin{tabular}{l|cc}
\hline
\textbf{dataset}    & \textbf{\# of triples} & \textbf{\# of rules} \\ \hline
FB15k-237           & 310,116                & 6,320                \\ 
FB-CVT-REV          & 125,124,274            & 14,355               \\
FB+CVT-REV          & 134,213,735            & 2,965                \\ 
ogbl-biokg          & 5,088,434              & 145,114              \\ \hline
\end{tabular}
}
\end{table}
\vspace{-.9 cm}
\begin{table}[ht]
\centering
\setlength{\tabcolsep}{1.3pt}
\caption{Evaluation results on the annotated data in phase 1}\label{table:results}
\vspace{-.2 cm}
\scalebox{0.9}{
\begin{tabular}{l|cccc|ccc|c}
\hline
\textbf{}            & \textbf{m ent$^\downarrow$} & \textbf{m rel$^\downarrow$} & \textbf{h ent$^\downarrow$} & \textbf{h rel$^\downarrow$} & \textbf{correctness$^\uparrow$} & \textbf{clarity$^\uparrow$} & \textbf{logical$^\uparrow$} & \textbf{perplexity$^\downarrow$} \\ \hline
\textbf{all}         & 0.10 & 0.04 & 0.29 & 0.10 & 4.36 & 4.67 & 2.36 & 36.14 \\ 
\textbf{prompt 1}    & 0.14 & 0.05 & 0.25 & 0.07 & 4.40 & 4.69 & 2.29 & 37.85 \\ 
\textbf{prompt 2}    & 0.06 & 0.03 & 0.34 & 0.12 & 4.32 & 4.64 & 2.44 & 34.36 \\ 
\textbf{unanimous}   & 0.13 & 0.03 & 0.35 & 0.12 & 4.34 & 4.68 & 2.29 & 33.80 \\ 
\textbf{majority}    & 0.08 & 0.05 & 0.24 & 0.07 & 4.37 & 4.66 & 2.43 & 38.30 \\ 
\hline
\end{tabular}
}
\end{table}
\vspace{-.9 cm}
\section{Experiments}\label{sec:experiments}
To mine the rules, we used the default settings of AMIE for optimized performance, with minimum thresholds of 0.1 for \textit{head coverage} and \textit{standard confidence}, and a maximum threshold of 3 for the number of atoms. The remainder of this section describes the datasets used in our study, along with the configuration details of our fine-tuning process.

\subsection{Datasets}\label{sec:data}
For our experiments, we leveraged four datasets. FB15k-237, a small subset of the Freebase dataset, was selected as it is a widely used benchmark for KG completion, recognized for avoiding the data leakage issues of FB15k~\cite{bordes2013translating}. Its multi-domain coverage makes it well-suited for extracting logical rules with diverse relations. As mentioned in Section~\ref{sec:prompt}, this dataset was only used in the first phase of prompt engineering to assess
how in-context examples affect explanation quality and establish baseline performance. FB-CVT-REV and FB+CVT-REV~\cite{shirvani2023comprehensive} datasets (Statistics shown in Table~\ref{tab:datastat}) are large-scale variants of the Freebase dataset designed to eliminate the data leakage issue previously identified in FB15k. FB+CVT-REV includes mediator entities (i.e., Compound Value Type nodes) originally present in Freebase to represent n-ary relations. In contrast, FB-CVT-REV converts n-ary relationships centered on a CVT node into binary relations by concatenating the edges that connect entities through the CVT node, a method also used in FB15k-237. As shown in Table~\ref{tab:datastat}, the conversion process has resulted in a higher number of rules in these two datasets compared to those in FB+CVT-REV. Including these datasets facilitates the analysis of large-scale data and the effects of mediator nodes and concatenated relationships on the derived rules and generated explanations.

The label of a concatenated relation is formed by merging the labels of two underlying relations. As a result, the label becomes lengthy, taking the format of \rel{domain1/type1/label1-/domain2/type2/label2}. Notably, the domains and even types can sometimes be identical in concatenated labels, but label1 and label2 are always distinct. This format differs from the simpler structure of standard relations, which follow the format of \rel{domain/type/label}. Thus, this added complexity can pose a greater challenge for LLMs in generating natural language explanations. For instance, consider the triple \triple{Dallas Cowboys}{/american\_football/game\_passing\_statistics/team-/american\_football/game\_passing\_statistics/player}{Tony Romo}. Following the aforementioned format, this relation indicates that \entity{Dallas Cowboys} and \entity{Tony Romo} participated in an n-ary relationship involving additional entities, with this property representing the result of converting the n-ary relationship to the binary format.

The fourth dataset used in our experiments is ogbl-biokg~\cite{hu2020open}, a well-established benchmark for KG completion tasks. This dataset was selected due to its domain-specific nature. As outlined in the motivation for this work, a key objective is to make prediction rules more accessible to non-experts and domain scientists. Using a domain-specific dataset such as ogbl-biokg allows us to better evaluate the applicability and effectiveness of our approach.

The statistics of the ogbl-biokg dataset are presented in Table~\ref{tab:datastat}. This dataset includes five types of entities: diseases, proteins, drugs, side effects, and protein functions. In total, it contains 51 distinct property types. Each entity in the dataset is identified by an ID that begins with its entity type—for example, "drug\_742". Unlike the Freebase dataset, in which entities may have multiple types, each entity in ogbl-biokg is associated with a single, unique type. This consistent naming convention was leveraged to infer the types of variable entities in rules, by examining rule instances as described in Section~\ref{sec:type}.

\subsection{Fine-Tuning Implementation Details}\label{sec:finetune}
As described in Section~\ref{sec:groundtruth} and Section~\ref{sec:eval-judge}, we construct ground-truth data for fine-tuning an LLM by combining a limited set of human-annotated examples with an LLM-as-a-judge framework. We applied this methodology to data from the FB-CVT-REV, FB+CVT-REV, and ogbl-biokg datasets, resulting in the creation of two ground-truth datasets—one for Freebase and one for ogbl-biokg. Each dataset consists of 500 rule–explanation pairs, split into 400 for training, 50 for validation, and 50 for testing. The 100 examples that were directly annotated by human evaluators were used exclusively for the validation and test sets. The datasets developed in this work are publicly available via our GitHub repository.

We selected Zephyr-7B-$\beta$, an instruction-tuned open-source language model, as the model for fine-tuning. All experiments were conducted using a single NVIDIA H100 80GB GPU. To accommodate GPU memory constraints, we applied 4-bit quantization during fine-tuning. The learning rate was set to 5e-5, providing a balance between convergence speed and training stability. We used a batch size of 2 for both training and evaluation, and both models—trained on the Freebase and ogbl-biokg datasets—were fine-tuned for two epochs.

\vspace{-.6 cm}
\begin{table*}[ht]
\centering
\setlength{\tabcolsep}{1.3pt}
\caption{Evaluation results on the annotated data in phase 2}\label{table:results2}
\vspace{-.2 cm}
\scalebox{0.88}{
\begin{tabular}{l|c|cccc|cc|c}
\hline
& \textbf{logical$^\uparrow$} & \textbf{m ent$^\downarrow$} & \textbf{m rel$^\downarrow$} & \textbf{h ent$^\downarrow$} & \textbf{h rel$^\downarrow$} & \textbf{correct$^\uparrow$} & \textbf{clarity$^\uparrow$} & \textbf{perplexity$^\downarrow$} \\
\hline
\multicolumn{9}{c}{\textbf{explanation from zero-shot prompt}} \\
\hline
\textbf{all}          & 2.58 & 0.06 & 0.10 & 0.22 & 0.09 & 3.94 & 4.12 & 29.05 \\
\textbf{2 atoms}      & 2.50 & 0.03 & 0.04 & 0.08 & 0.05 & 4.22 & 4.35 & 34.10 \\
\textbf{3 atoms}      & 2.62 & 0.08 & 0.13 & 0.31 & 0.12 & 3.78 & 3.99 & 26.21 \\
\textbf{binary}       & 2.59 & 0.08 & 0.10 & 0.18 & 0.08 & 4.04 & 4.22 & 31.02 \\
\textbf{mediator}     & 2.51 & 0.08 & 0.13 & 0.16 & 0.06 & 4.15 & 4.13 & 24.22 \\
\textbf{concatenated} & 2.60 & 0.02 & 0.08 & 0.35 & 0.15 & 3.63 & 3.91 & 27.63 \\
\hline
\multicolumn{9}{c}{\textbf{explanation from variable type prompt}} \\
\hline
\textbf{all}          & 2.58 & 0.05 & 0.07 & 0.21 & 0.13 & 4.21 & 4.19 & 33.07 \\
\textbf{2 atoms}      & 2.50 & 0.31 & 0.41 & 0.15 & 0.16 & 4.25 & 4.30 & 38.59 \\
\textbf{3 atoms}      & 2.62 & 0.07 & 0.08 & 0.24 & 0.11 & 4.18 & 4.12 & 29.97 \\
\textbf{binary}       & 2.59 & 0.06 & 0.03 & 0.20 & 0.11 & 4.32 & 4.28 & 34.11 \\
\textbf{mediator}     & 2.51 & 0.01 & 0.11 & 0.16 & 0.06 & 4.36 & 4.20 & 28.65 \\
\textbf{concatenated} & 2.60 & 0.05 & 0.11 & 0.25 & 0.20 & 3.88 & 3.99 & 33.33 \\
\hline
\end{tabular}}
\end{table*}
\vspace{-1.2 cm}
\begin{table*}[ht]
\centering
\setlength{\tabcolsep}{1.3pt}
\caption{Evaluation results on the annotated data in phase 3}\label{table:results3}
\vspace{-.2 cm}
\scalebox{0.78}{
\begin{tabular}{l|cc|c|cc|c|cc|c}
\hline
& \multicolumn{3}{c|}{\textbf{GPT-3.5 Turbo}} & \multicolumn{3}{c|}{\textbf{GPT-4o mini}} & \multicolumn{3}{c}{\textbf{Gemini 2.0 Flash}}\\
\hline
\textbf{}   & \textbf{correct$^\uparrow$} & \textbf{clarity$^\uparrow$} &  \textbf{perplexity$^\downarrow$}  & \textbf{correct$^\uparrow$} & \textbf{clarity$^\uparrow$} & \textbf{perplexity$^\downarrow$} & \textbf{correct$^\uparrow$} & \textbf{clarity$^\uparrow$} & \textbf{perplexity$^\downarrow$} \\ 

\hline
\textbf{all}             & 4.28&	4.26&	32.40&	4.45	&4.53	&31.57	&4.67&4.70&	27.19	 \\ 
\textbf{2 atoms}        & 4.38 & 4.43 &34.08 &4.52 &4.62& 40.96 &4.80& 4.76 &29.98  \\ 
\textbf{3 atoms}        & 4.22 & 4.17 &31.46 & 4.42 & 4.51 &26.26 &4.61& 4.68& 25.62 \\ 
\textbf{binary}     & 4.40 & 4.42 & 34.58 & 4.50 & 4.58 & 33.52 & 4.70 & 4.71 & 27.77 \\ 
\textbf{mediator}        & 4.13 & 4.07 &26.26& 4.24 &4.49& 26.82 &4.69& 4.63& 26.92 \\ 
\textbf{concatenated}       & 4.10 & 4.07 & 31.57 & 4.50 & 4.51 & 30.38 & 4.63 & 4.75 & 26.19\\ 
\hline
\end{tabular}}
\end{table*}
\vspace{-.9 cm}
\section{Results}\label{sec:results}
\vspace{-.2 cm}
\textbf{Prompt Engineering Results} \hspace{0.1cm} In phase 1, annotators identified which explanation better captured the rule's semantics, preferring more naturally worded explanations when semantic accuracy was comparable. After selecting the better explanation, they rated it using our evaluation metrics described in Section~\ref{sec:eval}. We calculated averages only for the majority-selected explanations. Table~\ref{table:results} presents average measures for all annotated rules, separated by prompt type (zero-shot vs few-shot, denoted as prompt 1 and prompt 2, respectively) and agreement level (unanimous vs majority). The measures are abbreviated as m ent, m rel, h ent, h rel, correctness, clarity, logical, and perplexity, respectively. Results demonstrate that the model generates relatively accurate and clear explanations with low perplexity. Of 100 annotated sentences, 49 were assigned to few-shot explanations and 51 to zero-shot explanations, with annotators reaching unanimous agreement on 48\% of rules. Missed or hallucinated elements were negligible, with most hallucinations occurring in relation labels, particularly for concatenated relations where the model generates additional entities or relations to explain complex labels.

Table~\ref{table:results2} presents the results for phase 2, averaged across all annotators. Explanation 2, generated using the prompt including the variable type, consistently shows higher correctness and clarity across all categories, highlighting the importance of type information for model comprehension. Both explanation types have minimal missing entities and relations. However, explanation 2 also shows slightly higher hallucination rates and increased perplexity. Rules with three atoms and those involving concatenated relations generally receive lower correctness and clarity scores, likely due to their increased complexity. Interestingly, despite these lower scores, annotators rated the rules from these two categories as more logically coherent.

Given the negligible number of hallucinated and missing entities and relations, we evaluated the explanations in phase 3 using only correctness, clarity, and perplexity. Table~\ref{table:results3} presents the results. Overall, the models exhibit trends similar to those observed in phase 2. For example, all models perform better on shorter rules, particularly those with only two atoms, and achieve higher performance on rules involving only binary relations compared to those with concatenated ones. GPT-3.5 Turbo shows improved performance with CoT prompting compared to its performance using only variable entities. This improvement is consistent across all categories except for rules that include mediator nodes. GPT-4o mini is the second-best performing model and demonstrates relatively strong performance on rules containing at least one concatenated relation. Gemini 2.0 Flash demonstrates the best overall performance. Its explanations are the most concise. For example, given the rule \triple{?a}{/travel/accommodation/accommodation\_type}{Luxury Resort} $\Rightarrow$ \triple{?a}{/travel/accommodation/price\_range}{High end}, GPT-4o mini generated: "If an accommodation is a Luxury Resort, then it falls within the High end price range," whereas Gemini 2.0 Flash produced: "Luxury resorts are in the high-end price range." However, in rare instances, it includes remarks such as, “Note: This rule is likely flawed.” Notably, the lowest clarity scores across all models are observed for rules involving mediator nodes. Additionally, most models exhibit their highest perplexity on rules with only two atoms, which is somewhat unexpected given the simplicity of these rules.

\textbf{Variable Entity Type Inference Performance} \hspace{0.1cm} To evaluate the effectiveness of inferring variable entity types, we removed the type information of variable entities from the prompt—information that originally existed in the Freebase dataset—and performed a type inference task. Subsequently, we generated explanations for the same set of 100 rules using the best-performing model, Gemini 2.0 Flash, and asked annotators to assess their correctness. This experiment was conducted exclusively on the Freebase dataset due to its high diversity of entity types, whereas the ogbl-biokg dataset contains only five entity types. In the majority of cases, the inferred types were highly accurate, achieving an overall correctness score of 4.53.

However, in a few cases, the model inferred a type that was more specific (i.e., a subtype) based on the instances it was exposed to. For instance, consider the following rule: \triple{?b}{/sports/competitor\_competition\_relationship/competitors-
/sports/competitor\_competition\_relationship/competition}{?f} $\land$ \triple{?a}{/sports/multi\_event\_tournament/athletic\_performances-
/sports/competitor\_competition\_relationship/competition}{?f} $\Rightarrow$ \triple{?a}{/sports/multi\_event\_tournament/athletic\_performances-
/sports/competitor\_competition\_relationship/competitors}{?b}. Since the random instances provided to the model were exclusively related to tennis, the model inferred that the variable \entity{?b} corresponds to the type tennis player. However, the rule itself is more general, and the correct type for variable \entity{?b} is sports professional athlete (\entity{/sports/pro\_athlete}).

\textbf{LLM-as-a-Judge Performance} \hspace{0.1cm}
Since clarity represents a highly subjective metric, our analysis concentrated on correctness assessment in this experiment. The LLM judge received identical information to human annotators: the rule, an instance of the rule, the list of variable entity types, and the corresponding explanation. To evaluate inter-rater reliability between LLM judges and human annotators, we employed Spearman correlation for rank-order agreement and Krippendorff's Alpha for consensus accounting for chance agreement. Annotator scores were averaged across multiple individuals and represented as floating-point values, while LLM judge scores were similarly formatted due to triple evaluation of each explanation for consistency.
Our evaluation reveals a Spearman correlation of 0.69, suggesting reasonably strong rank-order agreement and indicating that LLMs and humans tend to identify the same explanations as relatively better or worse. However, Krippendorff's Alpha of 0.59 reflects moderate consensus when accounting for chance agreement and the ordinal nature of ratings. This pattern suggests that while LLMs and humans show substantial agreement on relative explanation quality rankings, they exhibit more variability in absolute scoring, which is typical when comparing automated and human evaluation systems. Notably, the LLM judge consistently identified imperfect explanations, demonstrating reliable detection of quality deficiencies. These findings indicate promising potential for leveraging LLMs in scalable evaluation frameworks and automated dataset generation for natural language explanation tasks.

\textbf{Zephyr Performance} \hspace{0.1cm}
To evaluate explanation generation quality, we employed three complementary automatic metrics: BLEU~\cite{papineni2002bleu}, ROUGE~\cite{lin2004rouge}, and METEOR~\cite{banerjee2005meteor}, each capturing different aspects of text quality. BLEU measures n-gram precision between generated and reference explanations, assessing surface-level similarity and fluency. ROUGE evaluates recall-oriented overlap, particularly useful for measuring content coverage and informativeness of explanations. METEOR provides a more sophisticated assessment by incorporating synonymy, stemming, and word order, offering a more nuanced evaluation of semantic similarity.

The fine-tuning results, presented in Table~\ref{table:finetune}, demonstrate substantial improvements in explanation generation quality across both datasets, with particularly pronounced gains on the ogbl-biokg dataset. On the Freebase dataset, fine-tuning yielded consistent improvements across all metrics. The ogbl-biokg dataset showed even more dramatic enhancements, with BLEU scores improving from .38 to .55, ROUGE exhibiting the most substantial gain from .02 to .78, and METEOR rising from .36 to .81. The exceptionally low baseline ROUGE score (.02) on ogbl-biokg suggests the base model struggled significantly with content overlap in biomedical explanations, while the fine-tuned model's performance (.78) indicates successful domain adaptation. These results demonstrate that domain-specific fine-tuning is particularly effective for specialized knowledge graphs like ogbl-biokg, where technical terminology and domain-specific reasoning patterns are crucial for generating coherent natural language explanations.

\vspace{-.7 cm}
\begin{table}[h]
\centering
\setlength{\tabcolsep}{10.3pt}
\caption{Zephyr performance on Freebase and ogbl-biokg datasets}\label{table:finetune}
\vspace{-.2 cm}
\scalebox{0.9}{
\begin{tabular}{l|cc|cc}
\hline
& \multicolumn{2}{c|}{\textbf{Freebase}} & \multicolumn{2}{c}{\textbf{ogbl-biokg}} \\
\hline
\textbf{Metric}   & \textbf{Base} & \textbf{fine-Tuned} &  \textbf{Base} & \textbf{fine-Tuned} \\ 

\hline
\textbf{BLEU$^\uparrow$}             & .48&	.71&	.38&	.55		 \\ 
\textbf{ROUGE$^\uparrow$}        & .10 & .33 &.02 & .78 \\ 
\textbf{METEOR$^\uparrow$}     & .44 & .66 & .36 & .81  \\ 
\hline
\end{tabular}}
\end{table}
\vspace{-1 cm}
\section{Conclusion \& Future Work}\label{sec:conclusion}
\vspace{-.1 cm}
We presented Rule2Text, a comprehensive framework for generating natural language explanations of logical rules extracted from knowledge graphs using large language models. Through systematic experimentation across multiple datasets and prompting strategies, we demonstrated that Chain-of-Thought prompting combined with variable entity type information yields the most accurate explanations, with Gemini 2.0 Flash achieving the best performance. Our LLM-as-a-judge framework shows promising agreement with human evaluators, enabling scalable evaluation and ground truth dataset construction. Fine-tuning results demonstrated substantial improvements, particularly in domain-specific contexts where ROUGE scores improved from 0.02 to 0.78. Future research directions include evaluating more complex rules beyond AMIE's capabilities and developing more sophisticated type inference mechanisms.

\begin{credits}
\subsubsection{\ackname}
This material is based upon work supported by the National Science Foundation under Grants 
TIP-2333834. We extend our gratitude to the Texas Advanced Computing Center (TACC) for providing computing resources for this work's experimentation.
\end{credits}


\end{document}